\documentclass[letterpaper]{article} 
\usepackage{aaai25}  
\usepackage{times}  
\usepackage{helvet}  
\usepackage{courier}  
\usepackage[hyphens]{url}  
\usepackage{graphicx} 
\usepackage{natbib}  
\usepackage{caption} 
\usepackage{algorithm}
\usepackage{algorithmic}

\usepackage{newfloat}
\usepackage{listings}
\DeclareCaptionStyle{ruled}{labelfont=normalfont,labelsep=colon,strut=off} 
\floatstyle{ruled}
\newfloat{listing}{tb}{lst}{}
\floatname{listing}{Listing}
\usepackage[utf8]{inputenc} 
\usepackage[T1]{fontenc}    
\usepackage{url}            
\usepackage{booktabs}       
\usepackage{amsfonts}       
\usepackage{nicefrac}       
\usepackage{microtype}      
\usepackage{xcolor}         
\usepackage{graphicx}
\usepackage{amsmath}
\usepackage{enumerate}
\usepackage{comment}
\usepackage{enumitem}
\newcommand{\algname}{\emph{VarDrop}}
\usepackage{multirow}

\usepackage{cleveref}
\crefformat{footnote}{#2\footnotemark[#1]#3}
\usepackage[toc, page, header]{appendix}
\usepackage{colortbl}
\usepackage{subcaption}

\usepackage{algorithm}
\usepackage{algorithmic}
\usepackage{amsfonts}

\usepackage{amsthm}
\newtheorem{theorem}{Theorem}[section]

\theoremstyle{definition}
\newtheorem{definition}[theorem]{Definition}

\title{\algname: Enhancing Training Efficiency by Reducing Variate Redundancy in Periodic Time Series Forecasting}
\author{
    Junhyeok Kang\textsuperscript{\rm 1},
    Yooju Shin\textsuperscript{\rm 2}, 
    Jae-Gil Lee\textsuperscript{\rm 2}\thanks{Jae-Gil Lee is the corresponding author.}
}
\affiliations{
    \textsuperscript{\rm 1}LG AI Research\\
    \textsuperscript{\rm 2}KAIST\\
    junhyeok.kang@lgresearch.ai, \{yooju.shin, jaegil\}@kaist.ac.kr
}

\begin{document}

\maketitle

\begin{abstract}
Variate tokenization, which independently embeds each variate as separate tokens, has achieved remarkable improvements in multivariate time series forecasting. However, employing self-attention with variate tokens incurs a quadratic computational cost with respect to the number of variates, thus limiting its training efficiency for large-scale applications. To address this issue, we propose \algname{}, a simple yet efficient strategy that reduces the token usage by omitting redundant variate tokens during training. \algname{} adaptively excludes redundant tokens within a given \emph{batch}, thereby reducing the number of tokens used for dot-product attention while preserving essential information. Specifically, we introduce $k$-dominant frequency hashing\,($k$-DFH), which utilizes the ranked dominant frequencies in the frequency domain as a hash value to efficiently group variate tokens exhibiting similar periodic behaviors. Then, only representative tokens in each group are sampled through stratified sampling. By performing sparse attention with these selected tokens, the computational cost of scaled dot-product attention is significantly alleviated. Experiments conducted on public benchmark datasets demonstrate that \algname{} outperforms existing efficient baselines. 
\end{abstract}

 \section{Introduction}
\label{sec:intro}


Transformers have demonstrated impressive performance in time series forecasting, primarily due to their attention mechanisms\,\cite{trirat2024universal, shin2024recurve, nie2023a, zhang2023crossformer, wu2021autoformer, zhou2021informer}. Traditionally, most methods have employed temporal tokenization, treating all variates at a given timestamp as a single token. However, recent studies reveal that \emph{variate tokenization}\textemdash where each variate is embedded as a separate token\textemdash outperforms temporal tokenization in capturing inter-variate dependencies thereby increasing forecasting accuracy\,\cite{liu2023itransformer}. 
Due to its broad applicability to Transformers, variate tokenization has been adopted in recent advancements in multivariate time series forecasting\,\cite{liu2024timer,wang2024timexer,han2024mcformer}.

Despite its advantages, the feasibility of variate tokenization in real-world applications is hindered by the increasing number of variates\,($N$). Many public benchmark datasets demonstrate the high dimensionality often encountered in real-world applications. For instance, the Electricity dataset includes 321 variates, each corresponding to a customer, and the Traffic dataset comprises 862 variates, each representing a sensor\,\cite{wu2021autoformer}. The large $N$ introduces inefficiency, as the computational cost of attention mechanisms increases quadratically with $N$. This inefficiency results in a superfluous carbon footprint during the model training process. In addition, the large $N$ can lead to overfitting and reduced attention performance by diluting important information\,\cite{peysakhovich2023attention, liu2024lost}. Therefore, reducing $N$ while retaining the essential information is critical for the success of variate tokenization.

\begin{figure}
    \centering
    \includegraphics[width=\linewidth]{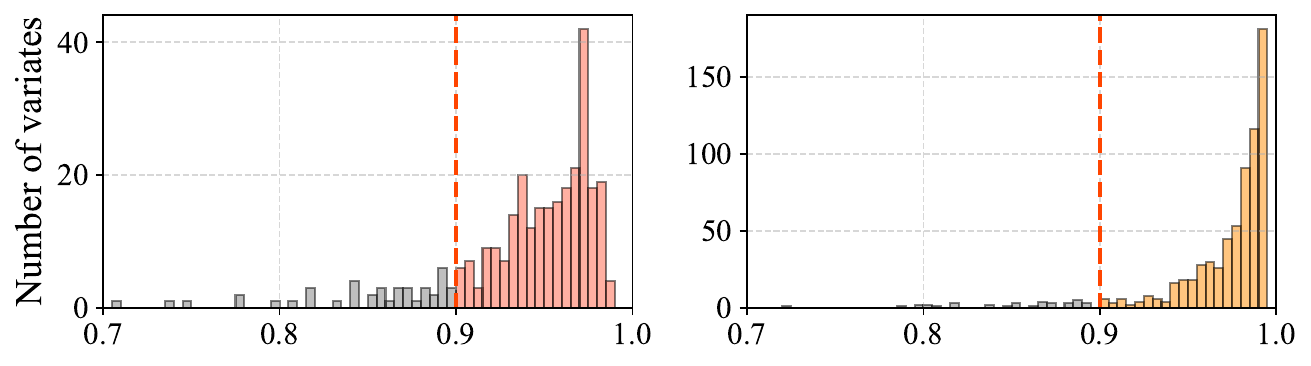}
    {\small (a) Electricity dataset.  \hspace{1.3cm} (b) Traffic dataset.}    
    \caption{\emph{Variate redundancy} in public periodic time series datasets. Most variates are highly correlated with others. For more detailed analysis, please refer to Appendix A.}
    \label{fig:max_correlation}
\end{figure}

To address this issue, we propose removing \emph{variate redundancy} in multivariate time series. As different variates often share the same characteristics such as trends and seasonality at some timestamps, excessive correlations between them are frequently observed in periodic time series. Figure~\ref{fig:max_correlation} shows the variate redundancy in the term of maximum Pearson correlation coefficient values (refer to Eq.~(4)) between the variates in two datasets.  
For the Electricity and Traffic datasets, 79.4\% and 94.7\% of all variates exhibit strong correlations with other variates exceeding 0.9, respectively. By selecting a few informative and distinguishing variates, we can significantly reduce computational cost while preserving the original periodic information of the data.


However, efficiently picking out the representative variates is challenging for two reasons. First, the correlation between variates fluctuates due to inherent distribution shifts in time series data. This correlation shift in sliding windows generates a unique set of correlations for each batch as shown in Appendix B. We should filter out redundant variates in an online manner due to this correlation shift. Second, computing similarity between variates requires extensive computational cost. Popular similarity metrics, such as the Pearson correlation coefficient and dynamic time warping\,(DTW), have computational complexity in the order of $O(N^2 T^2)$ where $T$ denotes the window length. In summary, we need a simple selection process with a fast similarity metric to leverage variate redundancy effectively.

To this end, we propose a simple yet efficient training strategy for variate tokens called \algname{}. Given a multivariate time series, \algname{} performs fast Fourier transform for each variate and identifies $k$-dominant frequencies whose amplitude values are top-$k$ in frequency domain. By ranking $k$ frequencies in the descending order of amplitude, $k$-dominant frequency hashing\,($k$-DFH) generates a meaningful hash for each variate. \algname{} groups the variates with the same hash and select representative variate tokens for each group through stratified random sampling. Scaled dot-product attention is then computed with selected tokens where each represents unique temporal patterns in each batch. As a result, \algname{} resolves variate redundancy and successfully reduces the number of tokens for each batch without incurring significant computational overhead.

Here we summarized our key contributions as follows: 
\begin{itemize}[leftmargin=10pt]
    \item We propose a simple yet efficient training strategy, \algname{}, which significantly improves computational efficiency by disregarding redundant variates in variate-tokenized Transformers.
    \item \algname{} leverages $k$-dominant frequency hashing that efficiently identifies similar tokens in each batch by applying the fast Fourier transform.
    \item We demonstrate the effectiveness of \algname{} through extensive experiments conducted on four benchmark datasets, comparing to state-of-the-art methods.
\end{itemize}

\section{Related Work}
\label{sec:related_work}

\subsection{Tokenization Strategies for Time Series Data}
Inspired by the success of Transformers in natural language processing, numerous Transformers have been proposed for multivariate time series forecasting. Previous studies, such as Autoformer\,\cite{wu2021autoformer}, Fedformer\,\cite{zhou2022fedformer}, Crossformer\,\cite{zhang2023crossformer}, and PatchTST\,\cite{nie2023a}, adopted a temporal tokenization method similar to language models, treating the values of all variates at a specific timestamp as a single token. The multiple variates are not considered individually, but processed as a whole when generating representations for each temporal token. As a result, these embedded temporal tokens fail to properly capture the correlations between different variates in multivariate time series. Moreover, temporal tokens could not consider relevant contexts due to the narrow receptive field\,\cite{shin2023cross}.

To address the limitations in temporal tokenization, iTransformer introduced variate tokenization\,\cite{liu2023itransformer}. It is a special type of tokenization as the various tokenization methods used in Flowformer\,\cite{wu2022flowformer} and treats each variate as one token, aiming to better model the variate dependencies. After the success of iTransformer, variate tokenization became a prevalent technique in forecasting models. Timer merges multiple variates from different domain into a single time series and treats the time series as a single token\,\cite{liu2024timer}. MCformer tokenizes each variate and mixes the variates to capture inter-variate correlations\,\cite{han2024mcformer}. TimeXer also leverages variate tokenization in the introduction of exogenous variates\,\cite{wang2024timexer}. 

\subsection{Efficient Transformers for Time Series Data}
Most previous efficient methods are designed for temporal tokens with sequential nature. Due to a number of tokens in the temporal axis, the sparse attention is prevalent to reduce the number of tokens in attention value computation. Here, a portion of query-key pairs is only considered instead of computing every query-key pair. LogSparse is one of the early methods for sparse attention, matching a query to the previous keys with an exponential step size and itself, showing the similar behavior in causal convolution\,\cite{li2019enhancing}. Big-Bird suggests a compound sparse attention method, containing global, local, and random query-key matching strategies\,\cite{zaheer2020big}. Reformer applies locality-sensitive hashing, which makes a chuck of similar tokens in an input sequence\,\cite{kitaev2020reformer}. These methods rely on the temporal locality of input sequences, making them unsuitable for variate tokens lacking a sequential nature.

There are also efficient Transformers not based on temporal locality between timestamps\,\cite{shin2022tclp}. Informer selects the dominant query-key pairs that has more influence in attention value computation\,\cite{zhou2021informer}. Autoformer adopts Fourier-based auto-correlation computation in the attention to reduce the computational complexity from $O(T^2)$ to $O(T\text{log}T)$\,\cite{wu2021autoformer}.
Pyraformer constructs a pyramidal graph in matching query-key pairs to scale the attention module into longer sequences\,\cite{liu2022pyraformer}. FEDformer randomly selects a fixed number of Fourier components in time series to have linear computational complexity in the forward pass\,\cite{zhou2022fedformer}. Crossformer proposes two-stage attention for temporal dimension and variate dimension to reduce the computation complexity from $O(N^2T^2)$ to $O(N^2T)$\,\cite{zhang2023crossformer}. However, these methods do not consider variate redundancy in multivariate time series for boosting efficiency in Transformers.
\section{Proposed Method: \algname{}}
\label{sec:method}

    \begin{figure*}[t]
        \centering
        \includegraphics[width=\linewidth]{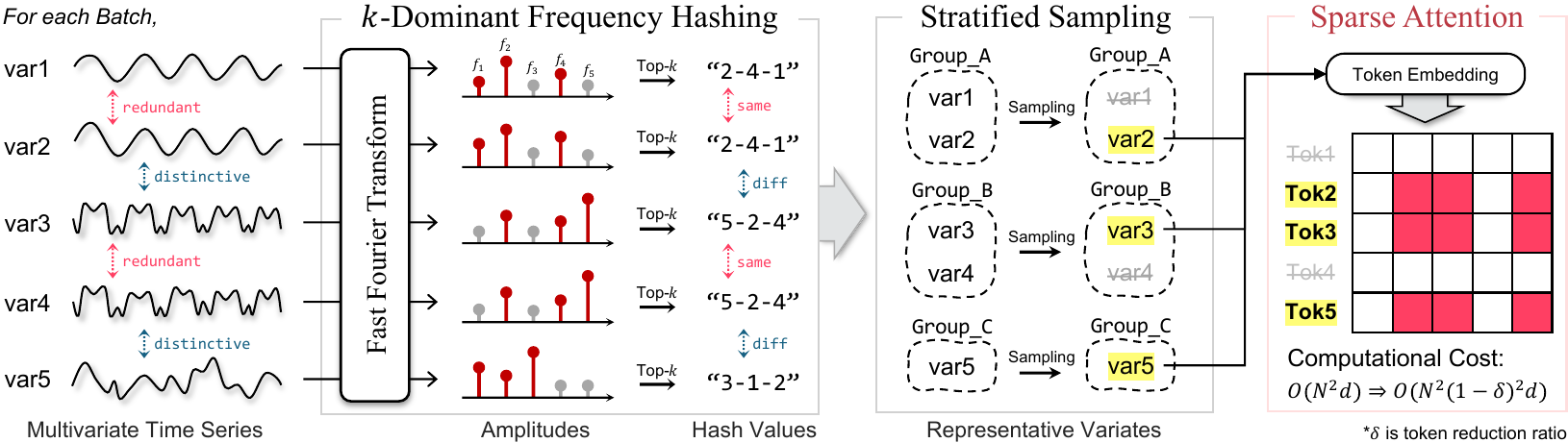}\\
        \caption{Overall procedure of \algname{}. Given a batch of multivariate time series, the hash values representing the top-$k$ amplitudes for each variate are generated through $k$-Dominant Frequency Hashing\,($k$-DFH). Then, stratified sampling is conducted based on groups of variates that share the same hash value to omit redundant variates. Finally, sparse attention is performed on the reduced set of variate tokens, enabling efficient training.}
        \label{fig:overview}
    \end{figure*}

\textbf{Problem Definition.} Given a multivariate periodic time series $\mathcal{X} \in \mathbb{R}^{N \times T}$, where $N$ is the number of variates and $T$ is the length of the time series, \textit{multivariate time series forecasting} aims to predict the forecast horizon $\mathcal{X}_{t+1:t+H} \in \mathbb{R}^{N \times H}$ based on a lookback window $\mathcal{X}_{t-T+1:t} \in \mathbb{R}^{N \times T}$. Variate tokenization converts a lookback window into the variate tokens $\mathcal{V} \in \mathbb{R}^{N \times d}$ through a embedding layer composed of multi-layer perceptron\,(MLP) shared by each variate. Note that $d$ is the dimension of the embedding. A variate-tokenized Transformer model then predicts the future values $\hat{\mathcal{X}}_{t+1:t+H} \in \mathbb{R}^{N \times H}$ for multivariate time series forecasting.

\subsection{Overview of \algname{}}
    
    Figure~\ref{fig:overview} illustrates the overall procedure of \algname{}. When a batch of multivariate periodic time series containing $N$ variates is given, $k$-dominant frequency hashing generates a hash value for each variate in the frequency domain, resulting in a total of $N$ hash values denoted as $\mathcal{H} \in \mathbb{R}^{N}$. These values represent the inherent periodic behaviors of each variate, enabling efficient grouping of variates with similar patterns. Some of the redundant variates that share the same patterns with others are then disregarded by stratified sampling. Using the selected $N(1-\delta)$ variate tokens, the computational complexity of self-attention is reduced from $O(N^2 d)$ to $O(N^2 (1-\delta)^2 d)$, where $d$ is the embedding size and $\delta$ is the token reduction ratio. Similar to Dropout\,\cite{srivastava2014dropout}, \algname{} is utilized only during the training stage, as multivariate time series forecasting requires predictions for all variables during the inference stage.

\subsection{Efficient Adaptive Variate Token Grouping}

    To achieve efficient sparse attention by disregarding redundant tokens, it is crucial to first identify groups of highly correlated variates. By leveraging the variate redundancy, we can selectively focus attention on the most significant tokens, thereby reducing computational complexity.

    \smallskip
    \noindent \textbf{$k$-Dominant Frequency Hashing.} We introduce $k$-dominant frequency hashing\,($k$-DFH), a simple solution that enables efficient grouping of variates in the frequency domain using the fast Fourier transform\,(FFT). To explain $k$-DFH, we first define the $k$-dominant frequency in Definition~\ref{def:dominant_frequency}.
    \begin{definition}[\sc $k$-Dominant Frequency] 
    \label{def:dominant_frequency}
        Given a time series $x$, after performing the Fourier transform, a frequency $f$ is \emph{dominant} if its corresponding amplitude spectrum $\mathcal{A}_f$ is among the top-$k$ amplitude values.
    \end{definition}
    According to Fourier theorem, a time series can be modeled with few Fourier spectra in the frequency domain\,\cite{brigham1988fast}. Thus, the overall periodic behaviors of variates can be successfully condensed as the proper $k$ dominant frequencies while ignoring unessential properties. $k$-DFH uses the order of these frequencies as a hash value, enabling efficient clustering of correlated variates.
    
    \begin{algorithm}[t]
        \caption{$k$-Dominant Frequency Hashing}
        \begin{algorithmic}[1]
            \small
            \REQUIRE A batch of multivariate time series data $\mathcal{B}$, the number of variates $N$, batch size $B$, the number of dominant frequencies $k$, a cutoff frequency $\varepsilon$.
            \ENSURE Hash values $\mathcal{H} \in \mathbb{R}^{N}$
            \STATE $\mathcal{A} \gets \texttt{Fast\_Fourier\_Transform}(\mathcal{B})$;
            \STATE $\hat{\mathcal{A}} \gets \texttt{Low\_Pass\_Filter}(\mathcal{A}, \varepsilon)$;
            \STATE $\bar{\mathcal{A}} \gets \frac{1}{|B|} \sum_{j=1}^{B}\hat{\mathcal{A}}_j$;
            \STATE $\mathcal{F}^* \gets \texttt{Top-k\_Frequency}(\bar{\mathcal{A}}, k)$; 
            \STATE $\mathcal{H} \gets \texttt{Generate\_Hash\_Value}(\mathcal{F}^*)$;
            \STATE $\textbf{return } \mathcal{H} \in \mathbb{R}^{N}$
        \end{algorithmic}
        \label{alg:pseudo_code}
    \end{algorithm} 
    
    \smallskip \noindent \textbf{Procedure of $k$-DFH.} 
    Algorithm \ref{alg:pseudo_code} outlines the $k$-DFH procedure. Consider a batch of time series $\mathcal{B}=\{\mathcal{X}_1,\dots,\mathcal{X}_B\}\ \in \mathbb{R}^{B \times N \times T}$, where $B$ represents the batch size. First, each variate $\mathcal{X}^{(i)} \in \mathbb{R}^{T}$ in the time domain is converted to the frequency domain through Fast Fourier Transform and real amplitudes $\mathcal{A}$ are only retained\,(Line 1). To support efficient operations, the frequency components above a specified cut-off frequency $\varepsilon$ are removed by incorporating a low-pass filter\,(LPF) while preserving low-frequency information\,(Line 2). Then, the amplitude values are averaged across instances within the batch to support batch-wise grouping, thereby alleviating instance-level noise\,(Line 3). After that, the $k$ dominant frequencies $\mathcal{F}^* \in \mathbb{R}^k$ are identified based on these amplitude values $\bar{\mathcal{A}}$\,(Line 4). Finally, $k$ dominant frequencies $\mathcal{F}^*$ are utilized as the input of hash function and generates hash value $\mathcal{H} \in \mathbb{R}^{N}$ for variate tokens\,(Line 5). Based on the hash values, the variates with the same hash value form a group, leading to multiple distinct groups that represent unique periodic behaviors, as shown in Figure~\ref{fig:overview}.

    \smallskip \noindent \textbf{Effect of Hyperparameters.}
    The $k$-DFH algorithm has a hyperparameter $k$ that determines the granularity of groups. Increasing $k$ results in a more fine-grained grouping, which reduces diversity between variates within groups and enhances the their similarity. Conversely, decreasing the value of $k$ leads to a coarse-grained grouping approach, resulting in larger differences in periodicity between groups and diminished intra-group similarity. The choice of $k$ is thus crucial and depends on the specific characteristics of the time series data and the desired balance between granularity and variability. We provide the experimental results concerning the impact of the $k$ value on the forecasting results and token reduction ratios in Section\,\ref{sec:sensitivity_analysis}. Because there is a trade-off between performance degradation and efficiency, the optimal selection of two hyperparameters depends on the application requirement. We also report its empirical evidence in Figure~\ref{fig:sensitivity_k_gs_ECL} of Section~\ref{sec:experiments}. For another minor hyperparameter $\varepsilon$ of LPF, by selecting an appropriate cut-off frequency, \algname{} enables more efficient and robust grouping by enhancing the representativeness of hash values while keeping essential information\,\cite{xu2024fits}.

    \smallskip \noindent \textbf{Time Complexity Analysis.} The $k$-DFH algorithm comprises steps including fast Fourier transform, low-pass filtering, amplitude averaging, identification of $k$ dominant frequencies, and hash value generation. The overall time complexity is dominated by the fast Fourier transform step, which has a complexity of $O(B \cdot N \cdot T \log T)$. Subsequent steps, involving low-pass filtering and amplitude averaging, have lower complexities of $O(B \cdot N \cdot \varepsilon)$ each. Finding the $k$ dominant frequencies among the averaged amplitude values $\bar{\mathcal{A}}$ requires sorting algorithm. Using an efficient sorting algorithm, the complexity is $O(N \cdot \varepsilon \log \varepsilon)$ for this step as there are $\varepsilon$ frequencies in $\bar{\mathcal{A}}$. Therefore, the overall time complexity of the $k$-DFH algorithm becomes $O(B \cdot N \cdot T \log T)$, making the algorithm efficient for datasets with large $N$. In Section~\ref{sec:efficiency_of_proposed_method}, we empirically verify that the runtime of the proposed method is significantly faster than existing efficient methods, demonstrating the practicality of $k$-DFH.
    
    \smallskip \noindent \textbf{Theoretical Analysis.} 
    The rationale behind $k$-DFH is that variates exhibiting similar periodic behaviors have the same dominant frequencies in the frequency domain. To provide its theoretical justification, we formalize this in Theorem~\ref{theorem:error_k_DFH_approx} where the \emph{proof} is provided in Appendix C. 

    \begin{theorem}[\sc Error of $k$-DFH Approximation]
        \label{theorem:error_k_DFH_approx}
        The error between a time series and its reconstructed signal from its hash value through $k$-DFH is given by the cumulative contribution of the non-dominant frequencies.
    \end{theorem}
    
    According to Theorem~\ref{theorem:error_k_DFH_approx}, the $k$ dominant frequencies capture the majority of the signal’s energy, and the reconstruction error using only these frequency components remains relatively small. This ensures that the errors between variates sharing the same hash value do not significantly deviate from one another. To support empirical evidence, we include visualizations of the variate groups generated by $k$-DFH in Section~\ref{sec:qualitative_analysis_visualization}. These visualizations illustrate that the variates within the same group exhibit similar periodic behavior, consistent with the theoretical expectations.

    \smallskip \noindent \textbf{Noise Robustness.} 
    One of the key advantages of $k$-DFH is its robustness to noise. Let $\mathcal{X}(t)$ be a time series composed of trends $\mathrm{T}(t)$, seasonality $\mathrm{S}(t)$ and noise $\mathrm{E}(t)$, such that $\mathcal{X} = \mathrm{T}(t) + \mathrm{S}(t) + \mathrm{E}(t)$. The frequency spectrum $\Phi(\mathcal{X})$ can be expressed as the sum of the frequency spectrums of the signal and noise: $\Phi(\mathcal{X}) = \Phi(\mathrm{T}) + \Phi(\mathrm{S}) + \Phi(\mathrm{E})$. Noise $\mathrm{E}(t)$ typically manifests as lower-amplitude components spread across the frequency spectrum, while the trends $\mathrm{T}(t)$ and seasonality $\mathrm{S}(t)$ contributes higher-amplitude components at specific frequencies. Since $\Phi(\mathrm{E})$ contributes relatively low amplitudes, the dominant frequencies below cutoff frequency $\varepsilon$ are predominantly those of $\mathrm{T}(t)$ and $\mathrm{S}(t)$, making $k$-DFH invariant to undesirable high-frequency noise.

\subsection{Sparse Attention via Stratified Sampling} 
\label{sec:sparse_attention_via_stratified_sampling}

    \noindent \textbf{Variate Reduction using Stratified Sampling.} By leveraging the $k$-DFH, similar variates can be grouped based on their dominant frequencies in the frequency domain. Once these groups are formed, stratified sampling is applied to selectively retain a subset of variates within each group, thereby eliminating redundant variate tokens. Formally, let $\mathcal{G}_i$ denote the set of variates in the $i$-th group, and let $gs$, a hyperparameter representing the group size, determine the maximum number of variates to retain per group. The retained subset of variates from group $\mathcal{G}_i$, denoted as $\mathcal{S}_i$, is defined as
    \begin{equation}
        \mathcal{S}_i \subseteq \mathcal{G}_i \quad \text{and} \quad |\mathcal{S}_i| = \min(|\mathcal{G}_i|, gs).
    \end{equation}
    The set of all variates retained across all groups, denoted as $\mathcal{S}$, is the union of the retained subsets from each group: 
    \begin{equation}
        \mathcal{S} = \cup_{i=1}^{G} \mathcal{S}_i, 
    \end{equation}
    where $G$ is the total number of generated groups. Our method is not limited to any specific sampling method during the stratified sampling process. Any sampling technique can be freely chosen based on the requirements of the application. In this study, we adopted random sampling as the sampling method due to its ease of implementation.

    \smallskip
    \noindent \textbf{Efficient Self-Attention Disregardng Redundant Variates Tokens.} In Transformers employing variate tokens, the role of self-attention mechanisms is capturing dependencies between variates\,\cite{liu2023itransformer, wang2024timexer}. The self-attention mechanism computes attention scores between each pair of variate tokens, resulting in a computational complexity of $O(N^2d)$, where $N$ is the number of tokens and $d$ is the embedding dimension. The attention scores are computed using the scaled dot-product attention, defined as:
    \begin{equation}
    \text{Attention}(Q, K, V) = \text{softmax}\left(\frac{QK^T}{\sqrt{d_k}}\right)V
    \end{equation}
    where $Q \in \mathbb{R}^{N \times d_k}$, $K \in \mathbb{R}^{N \times d_k}$, and $V \in \mathbb{R}^{N \times d_k}$ are the query, key, and value matrices, respectively, and $d_k$ is the dimension of the query, key, and value vectors. The bottleneck in the variate-tokenized Transformer lies in the quadratic computational complexity with respect to the number of tokens. As the number of tokens can be reduced by a token reduction ratio $\delta =1 - \frac{1}{N}\sum_{i}^G\min(|\mathcal{G}_i|, gs)$, the computational cost of self-attention decreases significantly, resulting in a complexity of $O(N^2 (1-\delta)^2 d)$. Moreover, the proposed method performs token reduction directly from raw variates, enhancing efficiency by eliminating the need for token embedding of redundant variates. Additionally, \algname{} is an architecture-free method that can be applied to any type of Transformer using various tokens. 

\begin{table*}[!ht]
\small
\centering

\begin{tabular}{c|c|cc|cc|cc|cc|cc|cc}
    
    \toprule
    
    \multicolumn{2}{c|}{Type} & \multicolumn{10}{c}{\sc Efficient Variate-Tokenized Transformers} & \multicolumn{2}{|c}{\sc Ground} \\[0.2em]
    
    \multicolumn{2}{c|}{Model} & \multicolumn{2}{c}{\algname{}} & \multicolumn{2}{c}{FlashAttention} & \multicolumn{2}{c}{Flowformer} & \multicolumn{2}{c}{Reformer} & \multicolumn{2}{c}{Informer} & \multicolumn{2}{|c}{iTransformer} \\
    
    \midrule
    \multicolumn{2}{c|}{Metric} & MSE & MAE & MSE & MAE & MSE & MAE & MSE & MAE & MSE & MAE & MSE & MAE \\[-.2em]
    
    \midrule
    \multirow{5}{*}{Electricity}
    & 96 & 0.153 & 0.245 & 0.178 & 0.265 & 0.183 & 0.267 & 0.182 & 0.275 & 0.190 & 0.286 & 0.150 & 0.242 \\
    & 192 & 0.167 & 0.257 & 0.189 & 0.276 & 0.192 & 0.277 & 0.192 & 0.286 & 0.201 & 0.297 & 0.166 & 0.256 \\
    & 336 & 0.183 & 0.275 & 0.207 & 0.294 & 0.210 & 0.295 & 0.210 & 0.304 & 0.218 & 0.315 & 0.184 & 0.276 \\
    & 720 & 0.220 & 0.305 & 0.251 & 0.329 & 0.255 & 0.332 & 0.249 & 0.339 & 0.255 & 0.347 & 0.214 & 0.302 \\
    \cline{2-14}\\[-0.8em]
    & \emph{Avg} & \textbf{0.181} & \textbf{0.271} & \underline{0.206} & \underline{0.291} & 0.210 & 0.293 & 0.208 & 0.301 & 0.216 & 0.311 & 0.178 & 0.269 \\[-.2em]
    
    \midrule
    \multirow{5}{*}{Traffic}
    & 96 & 0.396 & 0.274 & 0.464 & 0.320 & 0.493 & 0.339 & 0.617 & 0.356 & 0.632 & 0.367 & 0.398 & 0.272 \\
    & 192 & 0.417 & 0.281 & 0.479 & 0.326 & 0.506 & 0.345 & 0.629 & 0.361 & 0.641 & 0.370 & 0.418 & 0.279 \\
    & 336 & 0.435 & 0.289 & 0.501 & 0.337 & 0.526 & 0.355 & 0.648 & 0.370 & 0.663 & 0.379 & 0.431 & 0.286 \\
    & 720 & 0.472 & 0.308 & 0.524 & 0.350 & 0.572 & 0.381 & 0.694 & 0.394 & 0.713 & 0.405 & 0.465 & 0.304 \\
    \cline{2-14}\\[-0.8em]
    & \emph{Avg} & \textbf{0.430} & \textbf{0.288} & \underline{0.492} & \underline{0.333} & 0.524 & 0.355 & 0.647 & 0.370 & 0.662 & 0.380 & 0.428 & 0.285 \\[-.2em]
    
    \midrule
    \multirow{5}{*}{Weather}
    & 96 & 0.178 & 0.218 & 0.177 & 0.218 & 0.183 & 0.223 & 0.169 & 0.225 & 0.180 & 0.251 & 0.176 & 0.216 \\
    & 192 & 0.227 & 0.258 & 0.229 & 0.261 & 0.231 & 0.262 & 0.213 & 0.265 & 0.244 & 0.318 & 0.225 & 0.257 \\
    & 336 & 0.281 & 0.297 & 0.283 & 0.300 & 0.286 & 0.301 & 0.268 & 0.317 & 0.282 & 0.343 & 0.281 & 0.299 \\
    & 720 & 0.357 & 0.347 & 0.360 & 0.352 & 0.363 & 0.352 & 0.340 & 0.361 & 0.377 & 0.409 & 0.358 & 0.350 \\
    \cline{2-14}\\[-0.8em]
    & \emph{Avg} & \underline{0.261} & \textbf{0.280} & 0.262 & \underline{0.283} & 0.266 & 0.285 & \textbf{0.248} & 0.292 & 0.271 & 0.330 & 0.260 & 0.281 \\[-.2em]
    
    \midrule
    \multirow{5}{*}{Solar-Energy}
    & 96 & 0.202 & 0.238 & 0.205 & 0.237 & 0.208 & 0.239 & 0.203 & 0.241 & 0.198 & 0.237 & 0.205 & 0.236 \\
    & 192 & 0.237 & 0.262 & 0.239 & 0.263 & 0.244 & 0.266 & 0.236 & 0.266 & 0.232 & 0.263 & 0.239 & 0.263 \\
    & 336 & 0.254 & 0.275 & 0.250 & 0.275 & 0.258 & 0.277 & 0.249 & 0.276 & 0.248 & 0.276 & 0.249 & 0.273 \\
    & 720 & 0.252 & 0.274 & 0.252 & 0.277 & 0.259 & 0.279 & 0.253 & 0.279 & 0.251 & 0.278 & 0.250 & 0.275 \\
    \cline{2-14}\\[-0.8em]
    & \emph{Avg} & 0.236 & \textbf{0.262} & 0.236 & \underline{0.263} & 0.242 & 0.265 & \underline{0.235} & 0.265 & \textbf{0.232} & 0.264 & 0.236 & 0.262 \\[-.2em]
    
    \midrule
    \multicolumn{2}{c|}{\emph{Overall}} & \textbf{0.277} & \textbf{0.275} & \underline{0.299} & \underline{0.292} & 0.311 & 0.299 & 0.334 & 0.307 & 0.345 & 0.321 & 0.275 & 0.274 \\[-.2em]
    \midrule
    \multicolumn{2}{c|}{\emph{Relative error}\,(\%)} & \textbf{0.6} & \textbf{0.4} & \underline{8.6} & \underline{6.6} & 12.7 & 9.2 & 21.4 & 12.0 & 25.4 & 17.2 & \textbf{-} & \textbf{-} \\[-.2em]
    \bottomrule
    
\end{tabular}

\caption{Performance comparison of multivariate time series forecasting with four efficient baselines on the four benchmark datasets. {\sc Ground} refers to the ground-truth performance using all variate tokens. Variate tokenization is used for all compared methods. The forecasting horizon T $\in \{96, 192, 336, 720\}$ for all methods. The results are mostly taken from \citet{liu2023itransformer}.}

\label{table:compare_with_ground_truth.tex}
\end{table*}

\section{Experiments}
\label{sec:experiments}

In this section, we compare the proposed \algname{} with efficient Transformer baselines using public benchmark datasets for multivariate time series forecasting, evaluating both (i)\, forecasting performance and (ii)\,training efficiency to demonstrate \algname{}'s effectiveness.

\subsection{Experiment Settings}
\label{subsec:experiment_settings}

    \textbf{Datasets.}
    We conducted experiments on four real-world multivariate time series datasets, each containing a large number of variates: Electricity, Traffic, Weather, and Solar-Energy\,\cite{wu2021autoformer}. Further details on the data description are also provided in Appendix D.
    
    \noindent \textbf{Baselines.}
    We compare the performance of \algname{} with four efficient Transformers: Flowformer\,\cite{wu2022flowformer}, FlashAttention\,\cite{dao2022flashattention}, Reformer\,\cite{kitaev2020reformer} and Informer\,\cite{zhou2021informer}. FlashAttention reduces GPU memory bottlenecks by tiling vectors used in attention computation. All efficient baseline methods are modified to use variate tokenization following the previous study\,\cite{liu2023itransformer}.

    \noindent \textbf{Backbone Model and Ground Truth.} To objectively evaluate \algname{}, we adopted iTransformer\,\cite{liu2023itransformer}, a vanilla Transformer designed for the variate tokenization strategy, as our backbone model. We then used its dense attention results as the ground truth.

    \noindent \textbf{Evaluation Metrics.} For evaluation metrics, we choose the mean squared error\,(MSE) and the mean absolute error\,(MAE), consistent with previous work \cite{liu2023itransformer}.
    
    \noindent \textbf{Implementation Details.} Following previous studies\,\cite{liu2023itransformer}, we adopt same configuration for choosing optimization hyperparameters, such as the learning rate. In the experimental setup, the hyperparameters $k$ and $gs$ were selected from $\{3, 4\}$ and $\{5, 10, 20\}$, respectively. The source code is publicly available at \url{https://github.com/kaist-dmlab/VarDrop} and further details in Appendix E.
    

\subsection{Overall Performance Comparison}
    
    \textbf{Forecasting Performance.} 
        Table~\ref{table:compare_with_ground_truth.tex} summarizes the forecasting results of \algname{}, including ground-truth and efficient baselines across four real-world datasets. \algname{} demonstrates superior performance, achieving the lowest relative MSE and MAE, both less than 1\% compared to the ground truth. Please note that \algname{} accomplishes this high performance while utilizing a significantly reduced number of tokens for the attention process. \algname{} particularly excels in high-dimensional datasets such as Electricity and Traffic, increasing the feasibility of variate tokenization in large-scale applications. The high relative errors of efficient baselines stem from the improper handling of variate tokens, as they were designed for temporal tokens. To verify the robustness of \algname{}, we also provide the standard deviation of its performance across five independent runs initialized with different random seeds in Appendix F.

    \smallskip \noindent \textbf{Integration with Baselines.} 
        Due to its modular design, \algname{} can be easily attached to existing efficient methods. Specifically, by first applying \algname{} to reduce the number of tokens, the resulting reduced variates can be effectively used as input for existing Transformers. Table~\ref{table:collabo_efficient_baselines_with_ours.tex} presents the forecasting results of three efficient baselines and the corresponding results when integrated with \algname{}. This demonstrates that \algname{} further enhances the performance of efficient baselines by eliminating redundant variables.

    \def\arraystretch{1}
\begin{table}[t]
\small

\centering
\begin{tabular}{c|cccccc}

    \toprule
    
    Method & \multicolumn{2}{c}{Flowformer} & \multicolumn{2}{c}{Reformer} & \multicolumn{2}{c}{Informer} \\
    & \multicolumn{2}{c}{+\textbf{\algname{}}} & \multicolumn{2}{c}{+\textbf{\algname{}}} & \multicolumn{2}{c}{+\textbf{\algname{}}} \\
    
    Metric & MSE & MAE & MSE & MAE & MSE & MAE \\
    
    \midrule
    96 & 0.176 & 0.268 & 0.167 & 0.258 & 0.190 & 0.279 \\
    192 & 0.183 & 0.273 & 0.179 & 0.268 & 0.192 & 0.280 \\
    336 & 0.202 & 0.291 & 0.196 & 0.285 & 0.210 & 0.298 \\
    720 & 0.245 & 0.325 & 0.239 & 0.320 & 0.257 & 0.335 \\
    \midrule 
    \emph{Avg} & \textbf{0.202} & \textbf{0.289} & \textbf{0.195} & \textbf{0.283} & \textbf{0.212} & \textbf{0.298} \\
    \bottomrule

\end{tabular}

\caption{Forecasting results of existing efficient Transformers with \algname{} on the Electricity dataset.}

\label{table:collabo_efficient_baselines_with_ours.tex}
\end{table}

\subsection{Efficiency of \algname{}}
\label{sec:efficiency_of_proposed_method}

    \begin{table}[t]

    \small    
    \centering
    \setlength{\tabcolsep}{1.2mm}

    \begin{tabular}{lccc}
        \toprule
        Dataset & \# Used Tokens & \# Variates & Reduction Ratio \\
        \midrule
        Electricity  & 117.7{\footnotesize$\pm$3.8}  & 321 & \textbf{63.33\%} \\
        Traffic      & 188.4{\footnotesize$\pm$20.6} & 862 & \textbf{78.14\%} \\
        Weather      &   7.1{\footnotesize$\pm$1.3}  &  21 & \textbf{66.19\%} \\
        Solar-Energy &  20.0{\footnotesize$\pm$0.3}  & 137 & \textbf{85.38\%} \\
        \bottomrule
    \end{tabular}

    \caption{Token reduction results with standard deviations.} 

    \label{table:token_reduction_ratios_benchmark_datasets}
\end{table}
    \noindent \textbf{Token Reduction Results.} 
        Table~\ref{table:token_reduction_ratios_benchmark_datasets} presents the token reduction ratios achieved by \algname{} across four benchmark datasets: Electricity, Traffic, Weather, and Solar-Energy. Our method adaptively identifies and drops redundant variate tokens during the training stage for each batch. Therefore, we report the average number of tokens over all iterations within an epoch, along with the corresponding standard deviation. Table \ref{table:token_reduction_ratios_benchmark_datasets} verifies that \algname{} significantly reduces the number of tokens required for training. For example, in the Traffic dataset, \algname{} uses an average of 188.4$\pm$20.6 tokens out of 862, achieving a token reduction ratio of 78.14\%. Similarly, the Solar-Energy dataset exhibits an impressive reduction ratio of 85.38\%, with only 20.0$\pm$0.3 tokens used out of 137. These results demonstrate that \algname{} can be adopted for large-scale applications due to its efficiency and scalability.

    \smallskip \noindent \textbf{Comparison of Training Times.} 
        To validate the efficiency of \algname{}, we compared the average running time per iteration with efficient baselines during the training stage. Table~\ref{table:run_time_efficient_baselines} illustrates that \algname{} achieves the lowest average training time compared to the baselines. The ranking of training speeds for efficient baselines varied depending on the dataset. Notably, \algname{} increased the training speed of iTransformer from 68ms/iter to 33ms/iter, improving it by 2.06 times. 

        \begin{table}[t]
    
    \setlength{\tabcolsep}{1.6mm}
    \small
    \centering

    \begin{tabular}{l|cccc|c}
        \toprule
        Method&    Electricity  & Traffic      & Weather      & Solar & \emph{Avg}\\
        \midrule
        iTransformer & 40.9 & 198.5 & 19.4 & \textbf{13.1} & 68.0\\
        Informer     & 40.2 & \underline{118.9} & \underline{17.0} & 16.6 & \underline{48.2} \\
        Flowformer   & \underline{38.4} & 123.1 & 24.9 & 17.1 & 50.9 \\ 
        Reformer     & 84.7 & 308.2 & 21.0 & 27.0 & 110.2\\
        \textbf{\algname{}} & \textbf{30.8}  & \textbf{72.7}  & \textbf{12.6} & \underline{15.9}  &\textbf{33.0}\\
        \bottomrule
    \end{tabular}

    \caption{Comparison of training times on benchmark datasets with an input-96-predict-96 setting. The unit is \emph{ms/iteration}.}
    \label{table:run_time_efficient_baselines}

\end{table}

    


    \noindent \textbf{Comparison of Memory Footprints.} 
        We also report the results of comparing the GPU memory footprints of the baselines in Table~\ref{table:memory_footprint_efficient_baselines}. We observed that our proposed methodology uses less memory than all other efficient baselines. Remarkably, \algname{} utilizes 2.22 GB of memory, which is 65.1\% of the memory foorprint of iTransformer. This result is consistent with the token reduction ratio presented in Table~\ref{table:token_reduction_ratios_benchmark_datasets}. As discussed in Section~\ref{sec:sparse_attention_via_stratified_sampling}, this improved efficiency is attributed to the high reduction ratios, achieving $O(N^2 (1-\delta)^2 d)$. This significant reduction in computational overhead allows the Transformers exploiting variate tokens to reduce training time and memory usage. Overall, these results suggest that \algname{} is a promising approach for efficient training using variate tokenization.
    
        \begin{table}[t]
    \small

    \centering

    \begin{tabular}{lc}
        \toprule
        Method & Memory Footprint \\
        \midrule
        iTransformer & 3.41\,GB\\
        Informer     & 2.87\,GB\\
        Flowformer   & 3.33\,GB\\ 
        Reformer     & 4.12\,GB\\
        Flashformer  & \underline{2.87\,GB}\\
        \textbf{\algname{}} & \textbf{2.22\,GB}\\
        \bottomrule
    \end{tabular}

    \caption{Comparison of GPU memory footprints of efficient baselines on Electricity with an input-96-predict-96 setting.} 
    \label{table:memory_footprint_efficient_baselines}

\end{table}
    
    \vfill

\subsection{Qualitative Analysis through Visualization.}
\label{sec:qualitative_analysis_visualization}

\begin{figure}[!t]
            \centering
            \includegraphics[width=0.19\linewidth]{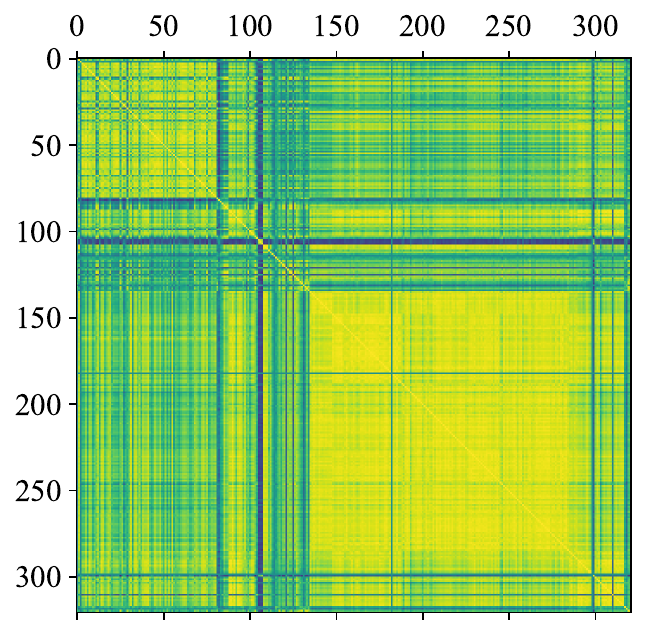}
            \hspace{0.05em}
            \includegraphics[width=0.19\linewidth]{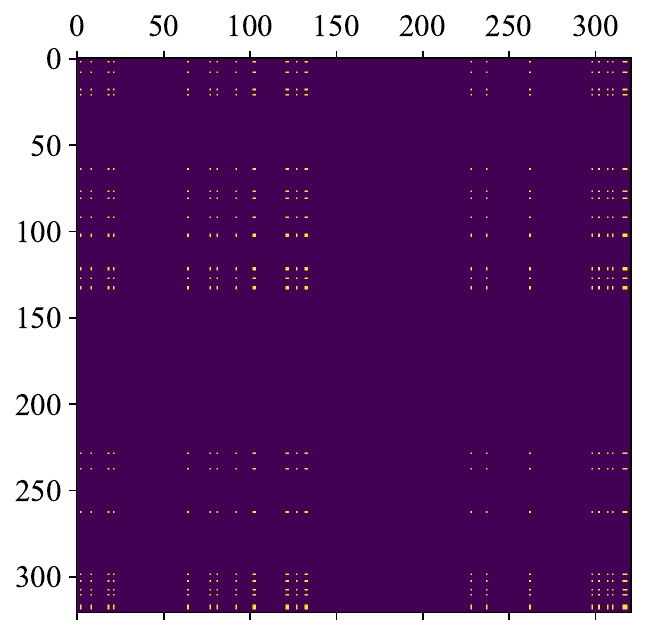}
            \includegraphics[width=0.19\linewidth]{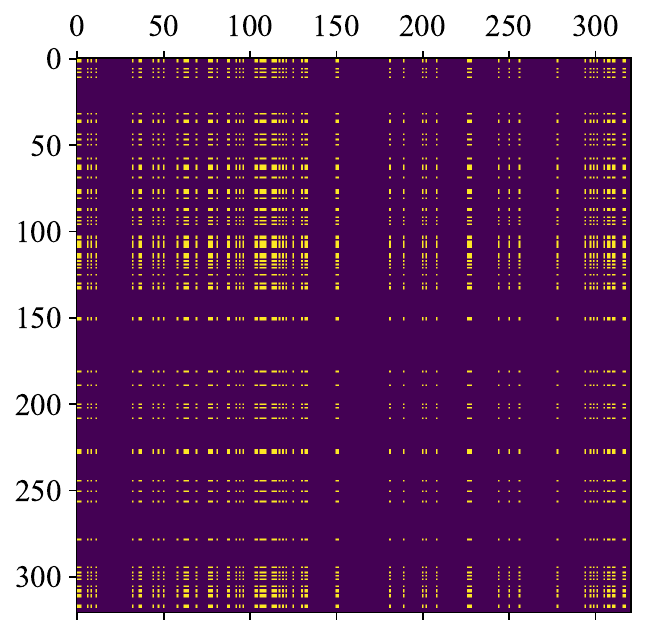}
            \includegraphics[width=0.19\linewidth]{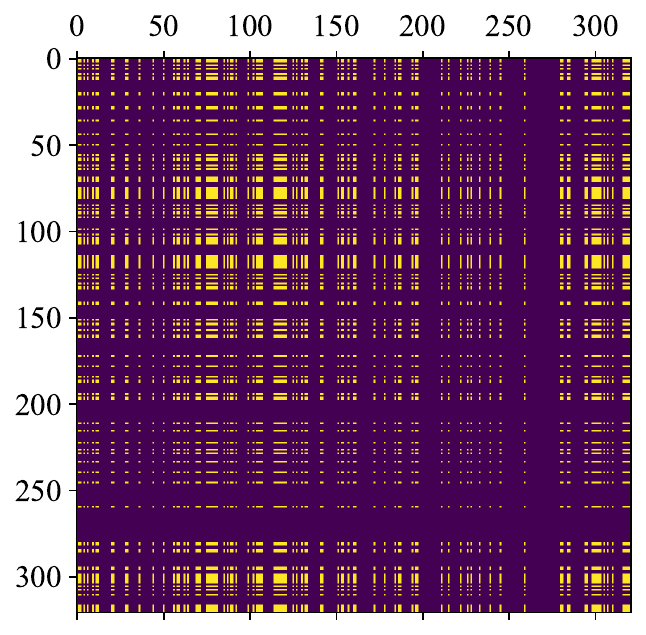}
            \includegraphics[width=0.19\linewidth]{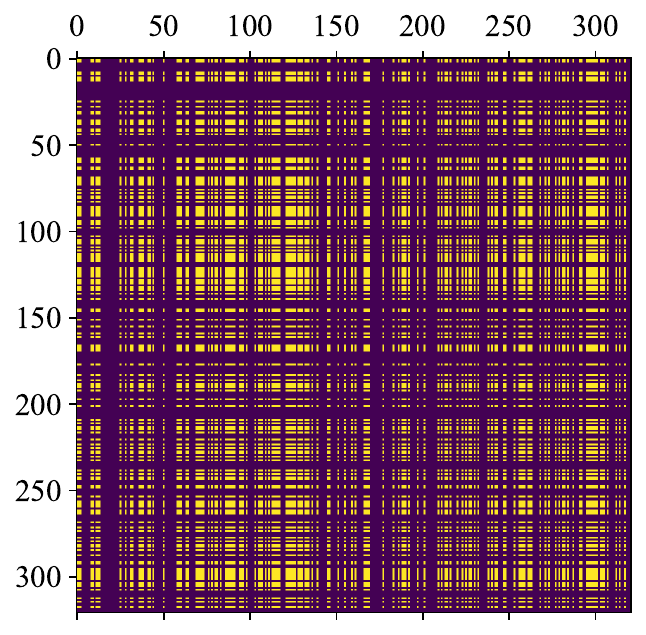}\\
            {\small (a) Correlation. \hspace{0.05cm} (b) $gs$=1. \hspace{0.25cm} (c) $gs$=5. \hspace{0.25cm} (d) $gs$=10. \hspace{0.2cm} (e) $gs$=20. \hspace{0.05cm} }\\
            \caption{Correlation matrix and corresponding sparse matrices with varying $gs$ values on the Electricity dataset.}
            \label{fig:sparse_matrices_varying_max_group_size}
        \end{figure}

    \noindent \textbf{Effects of Maximum Group Size.} 
        Increasing the group size $gs$ results in a higher density of the sparse matrix, leading to the lower level of token reduction ratio $\delta$. Figure~\ref{fig:sparse_matrices_varying_max_group_size} demonstrates how changes in the group size $gs$ affect sparse matrices on the Electricity dataset. The token reduction ratios $\delta$ for each matrix are as follows: (b) 93.1\%, (c) 76.1\%, (d) 63.5\%, and (e) 47.3\%. The redundant variates indicated in yellow in Figure~\ref{fig:sparse_matrices_varying_max_group_size}(a) are dropped less frequently even with large $gs$ values via stratified sampling . Please refer to Figure~10 of Appendix G for more visualization results on other benchmark datasets.
        
    \begin{figure*}[t]
        \centering
        \includegraphics[width=\linewidth]{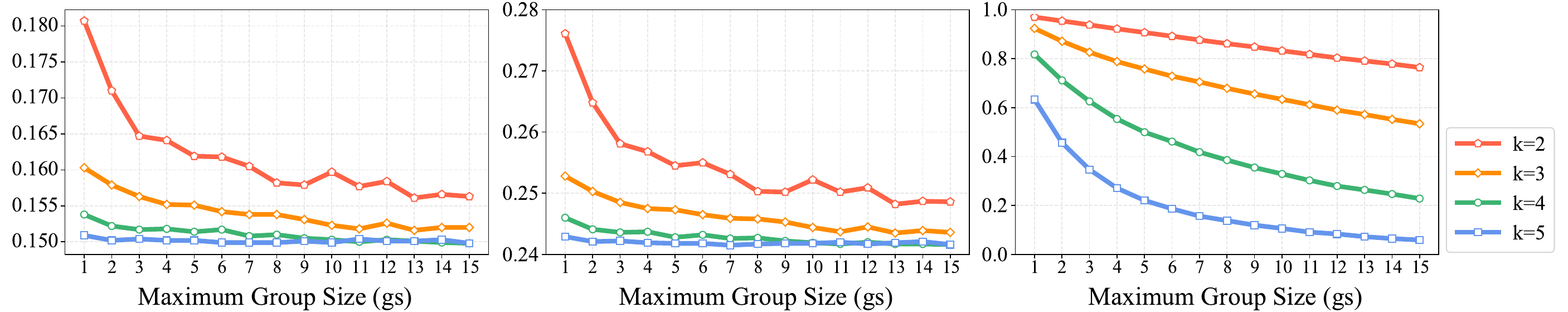}\\
        {\small  \hspace{.2cm} (a) MSE\,($\downarrow$). \hspace{3.6cm} (b) MAE\,($\downarrow$). \hspace{2.6cm} (c) Token Reduction Ratio $\delta$\,($\uparrow$).}\\
        \caption{Forecasting results with input-96-predict-96 setting with their token reduction ratios for varying two hyperparameters $k \in \{2,3,4,5\}$ and $gs \in \{1, \ldots, 15\}$ on the Electricity dataset.}
        \label{fig:sensitivity_k_gs_ECL}
    \end{figure*}

        \begin{figure}[t]
            \centering
            \includegraphics[width=0.9\linewidth]{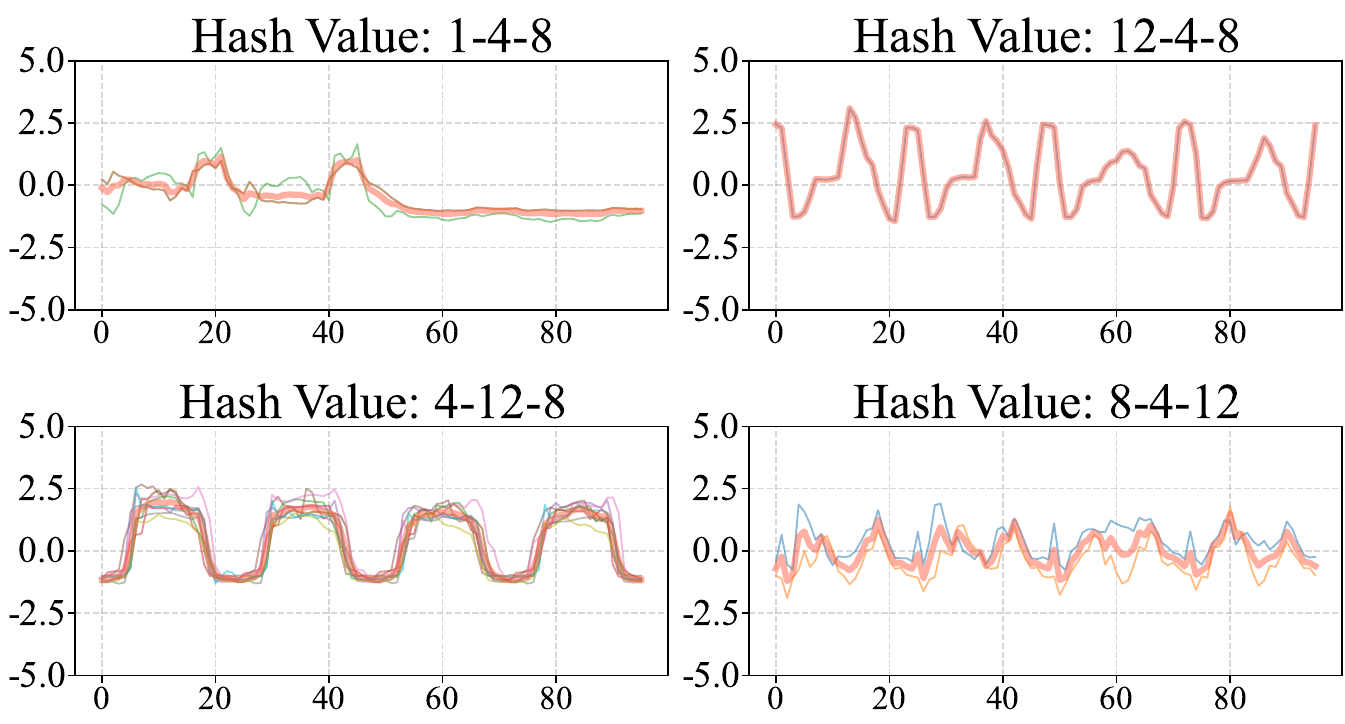}\\[-.3em]
            {\small (a) Electricity dataset\,($k$\,=\,3).}\\[.2em]
            
            \includegraphics[width=0.9\linewidth]{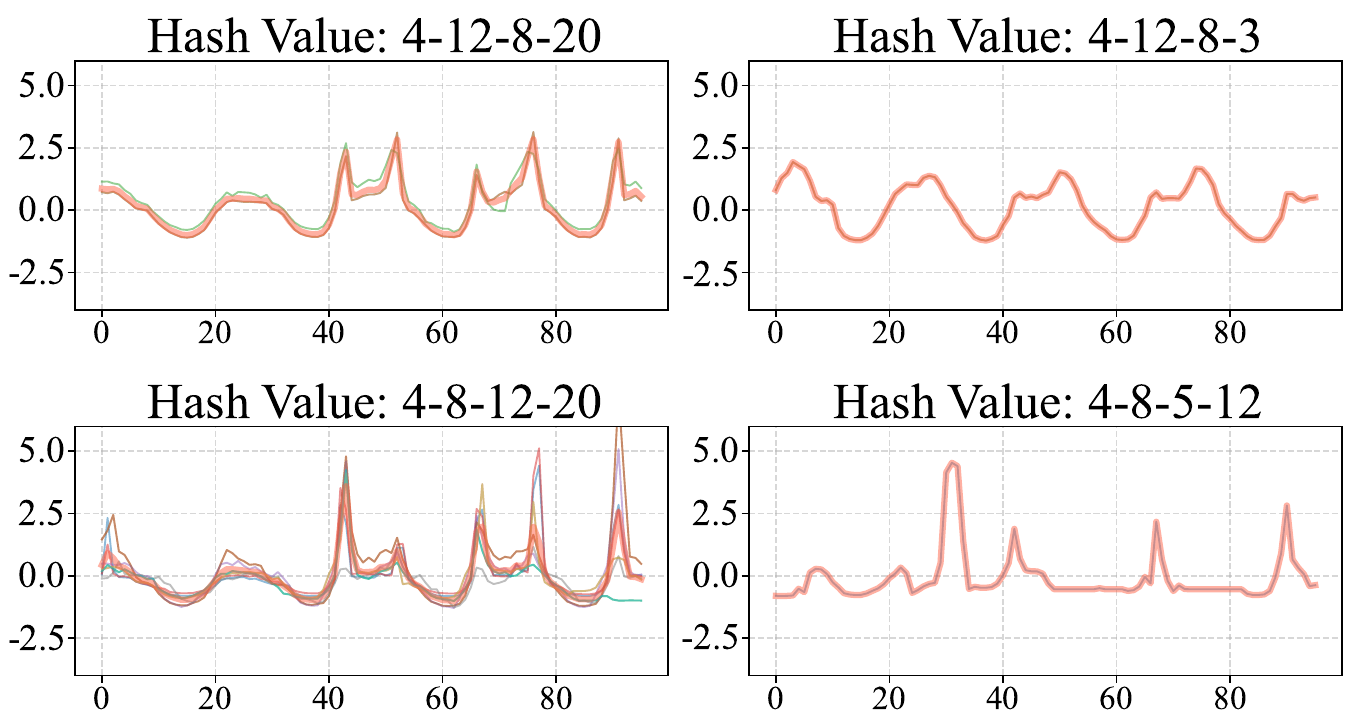}\\[-.3em]
            {\small (b) Traffic dataset\,($k$\,=\,4).}\\
            \caption{Visual examples of variates grouped by $k$-DFH. The red lines represent the mean values of the variates within each group, illustrating the distinctive periodic patterns. The hash values represent the ordered set of $k$ dominant frequencies.}
            \label{fig:examples_hash_group_main}
        \end{figure}

    \newpage \noindent \textbf{Visualization of $k$-DFH's Grouping Results.} 
        Figure~\ref{fig:examples_hash_group_main} illustrates various grouping results obtained from $k$-DFH with each hash value. Each subfigure demonstrates how the variates are clustered based on their hash values, representing overall periodic behaviors. In this visualization, the order of dominant frequencies distinguishes different patterns. In Figure~\ref{fig:examples_hash_group_main}(a), two variates with hash values 4-12-8 and 12-4-8 display distinct periodic behaviors despite having the same set of dominant frequencies. This verifies the effectiveness of $k$-DFH in grouping only similar variates together, as proved in Theorem~\ref{theorem:error_k_DFH_approx}. In Figure~\ref{fig:examples_hash_group_main}(b), the variate with the hash value 4-8-5-12 exhibits unique spikes that distinguish it from other samples, resulting in a different hash value.  This implies that the $k$-DFH method can effectively isolate variates that do not share common periodic characteristics with others. Overall, $k$-DFH efficiently assigns highly correlated variates to the same clusters. Please refer to Figure~9 for more results.

\subsection{Hyperparameter Sensitivity Analysis}
\label{sec:sensitivity_analysis}
    To examine the effect of \algname{}'s two crucial hyperparameters, $k$ and $gs$, we conducted a sensitivity analysis by varying these values. Figure~\ref{fig:sensitivity_k_gs_ECL} illustrates the two forecasting errors\,(MSE and MAE) and the token reduction ratio $\delta$ for different values of the length of hash value $k \in \{2, 3, 4, 5\}$ and the maximum group size $gs \in \{1, \ldots, 15\}$ on the Electricity dataset. As shown in these results, as the value of $k$ increases, the MSE and MAE decrease, while the token reduction ratio $\delta$ increases. 
    Interestingly, there are sweet spots in determining the proper levels of $k$ and $gs$. For instance, the forecasting errors and the reduction ratio when $k=4$ and $gs=1$ outperform those when $k=3$ and $gs=5$. This observation indicates that a higher reduction ratio does not necessarily result in lower forecasting performance and shows that \algname{}'s originality differs from random sampling. This is because, as $k$ increases, the groups generated by $k$-DFH can represent the underlying behaviors in the data with greater precision, as evidenced in Theorem~\ref{theorem:error_k_DFH_approx}. Consequently, selecting the appropriate levels of these two hyperparameters $k$ and $gs$ can significantly enhance both accuracy and computational efficiency, tailored to the specific needs of the application.

\vfill

\section{Conclusion}
\label{sec:conclusion}

This paper introduces a simple yet efficient training strategy for Transformers using variate tokenization, named \algname{}, for periodic time series forecasting. \algname{} adaptively identifies groups of variates that exhibit similar behaviors through $k$-DFH. The number of variate tokens is then reduced by disregarding redundant variates within each group via stratified sampling. By dropping these redundant variate tokens for each batch, the training efficiency of the attention mechanism is significantly enhanced. Experimental results on benchmark datasets demonstrate that the proposed method outperforms state-of-the-art efficient methods. Furthermore, due to its modularity, our approach can be easily applied to existing Transformers utilizing variate tokens. We hope that our work enhances the potential of variate tokenization in large-scale applications with numerous variables.

\section*{Acknowledgements}
This work was supported by Institute of Information \& Communications Technology Planning \& Evaluation\,(IITP) grant funded by the Korea government\,(MSIT) (No.\ RS-2020-II200862, DB4DL: High-Usability and Performance In-Memory Distributed DBMS for Deep Learning, 50\% and No.\ RS-2022-II220157, Robust, Fair, Extensible Data-Centric Continual Learning, 50\%).

\bibliography{6-Reference}
\clearpage 
\appendix



\section{Variate Redundancy in Multivariate Time Series Benchmark Datasets}
\label{sec:appendix:redundancy_datasets}

    In the multivariate time series, there are \emph{redundant} variates that exhibit high correlations with each other due to common underlying factors. This redundancy can hinder the efficiency of the attention mechanism, making it crucial to identify and eliminate these redundant variates. 
    
    To visualize the variate redundancy in multivariate time series, we use Pearson Correlation coefficients\,\cite{liu2023itransformer} of each variate of the input time series by the following equation:
    \begin{equation}
    \label{eq:pearson_corr_coef}
        \rho_{v,w}=\frac{\sum_i (v_i - \bar{v}) (w_i - \bar{w})}{\sqrt{\sum_i (v_i - \bar{v})^2} \sqrt{\sum_i (w_i - \bar{w})^2}},
    \end{equation}
    where $v_i$ and $w_i$ are the values of the two variates at the $i$-th time point, and $\bar{v}$ and $\bar{w}$ are the mean values of the variates $v$ and $w$, respectively. This coefficient measures the linear correlation between two variates, with values ranging from -1 to 1, where 1 indicates a perfect positive linear relationship, -1 indicates a perfect negative linear relationship, and 0 indicates no linear relationship.

    \begin{figure}[h]
        \centering
        \includegraphics[width=0.23\linewidth]{figures/correlations/corr_matrix_ECL.pdf}\hspace{0.1em}
        \includegraphics[width=0.23\linewidth]{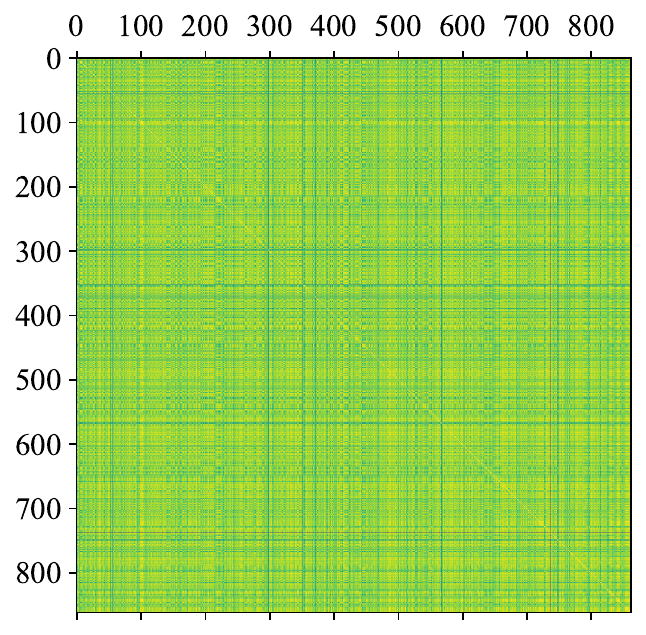}\hspace{0.1em}
        \includegraphics[width=0.23\linewidth]{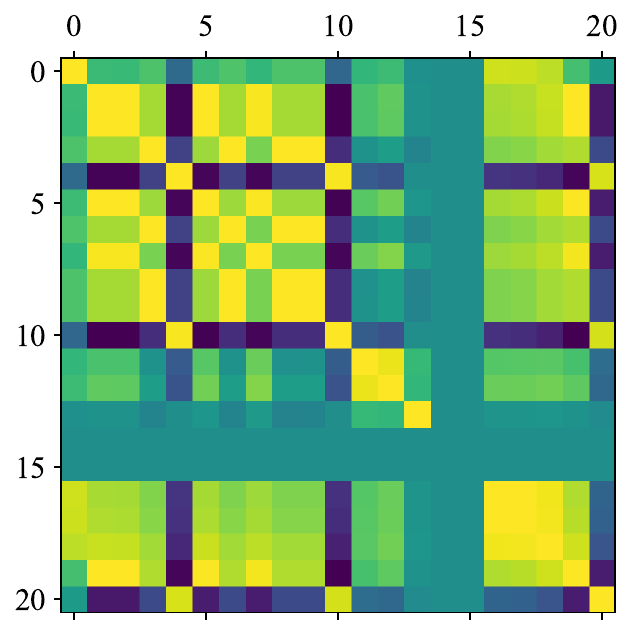}\hspace{0.1em}
        \includegraphics[width=0.23\linewidth]{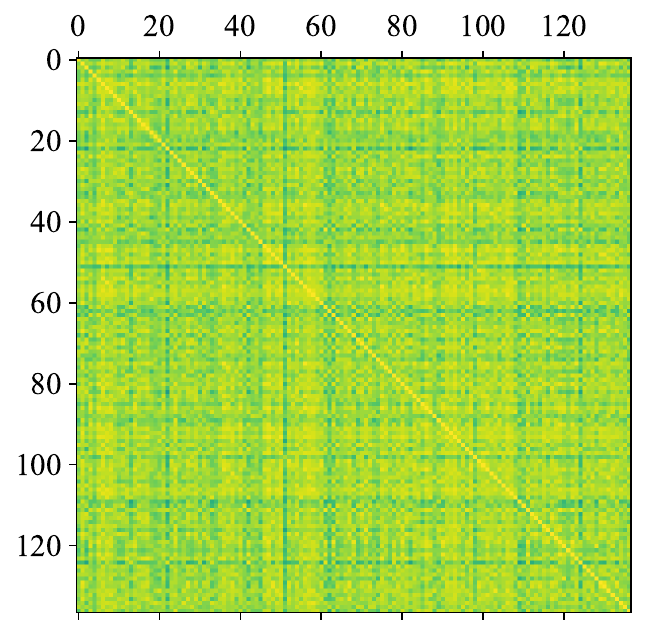}\hspace{0.1em}\\
        {\small \hspace{-.3cm} (a) Electricity. \hspace{.3cm} (b) Traffic. \hspace{.45cm} (c) Weather. \hspace{.55cm} (d) Solar.}\\[-.0em]
        \caption{Pearson correlation coefficient matrix of four benchmark datasets. Yellow colors represent strong correlations, while darker colors represent weaker correlations.}
        \label{fig:correlation_matrix_benchmark_dataset}
    \end{figure}
    
    Figure~\ref{fig:correlation_matrix_benchmark_dataset} displays the Pearson correlation coefficients between variates from the last 96-length sequence in the training data of four benchmark datasets: Electricity, Traffic, Weather, and Solar-Energy. This visualization demonstrates how groups of variates are influenced by shared factors. For example, in the Electricity dataset, distinct yellow blocks suggest that subsets of households exhibit similar patterns due to common influences such as weather conditions or local holidays. In the Traffic dataset, traffic volume measurements from various road segments within the same area are affected by rush hour patterns and local events. Similarly, in the Weather dataset, temperature and humidity measurements from nearby locations show high correlations due to regional climate patterns. 

    \begin{figure}[h]
        \centering
        \includegraphics[width=\linewidth]{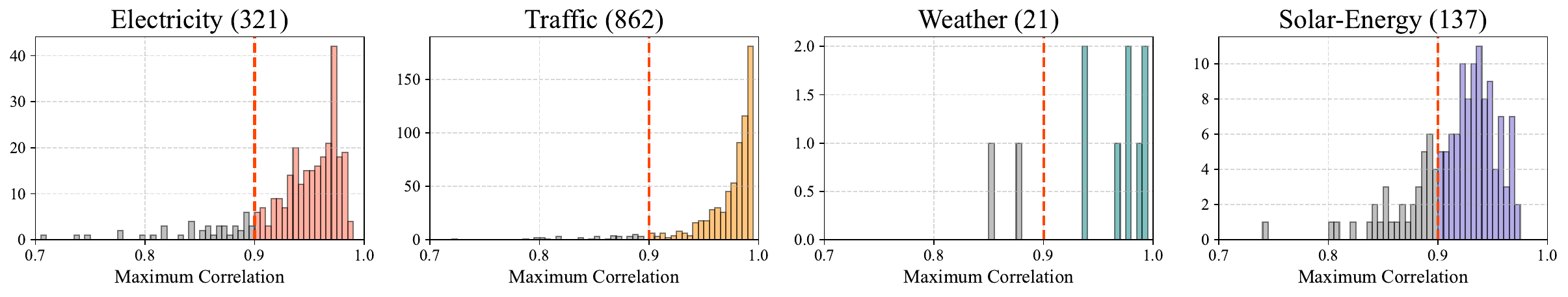}\\
        \vspace{-0.2cm}
        \caption{High correlations between variates in public benchmark datasets. The numbers in parentheses indicate the number of variates in each dataset.}
        \label{fig:max_correlation_hist_four_dataset}
    \end{figure}

    Figure~\ref{fig:max_correlation_hist_four_dataset} shows histograms of the maximum Pearson correlation coefficient values\,(refer to Eq.~\eqref{eq:pearson_corr_coef}) that each variate has with other variates in four benchmark datasets. For the Electricity, Traffic, Weather, and Solar-Energy datasets, the percentages of variates with a strong correlation of 0.9 or higher are 79.44\%, 94.66\%, 76.19\%, and 73.72\%, respectively. Such strong correlations between variates imply that very similar operations are being performed separately during the attention computation process for variate tokens. Therefore, alleviating variate redundancy can increase the efficiency of the Transformer by avoiding these overlapping computations.

\section{Correlation Shifts in Sliding Windows}
\label{sec:appendix:correlation_shift}
\begin{figure}[h]
        \centering
        \includegraphics[width=\linewidth]{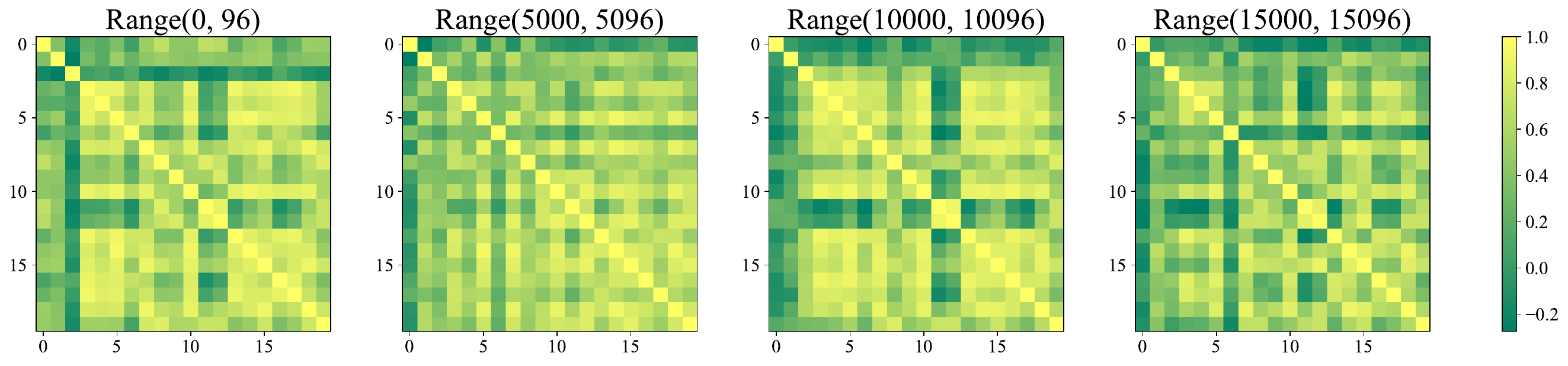}\\[-.2em]
        {\small (a) Correlation shifts in the Electricity dataset.}\\[.5em]
        \includegraphics[width=\linewidth]{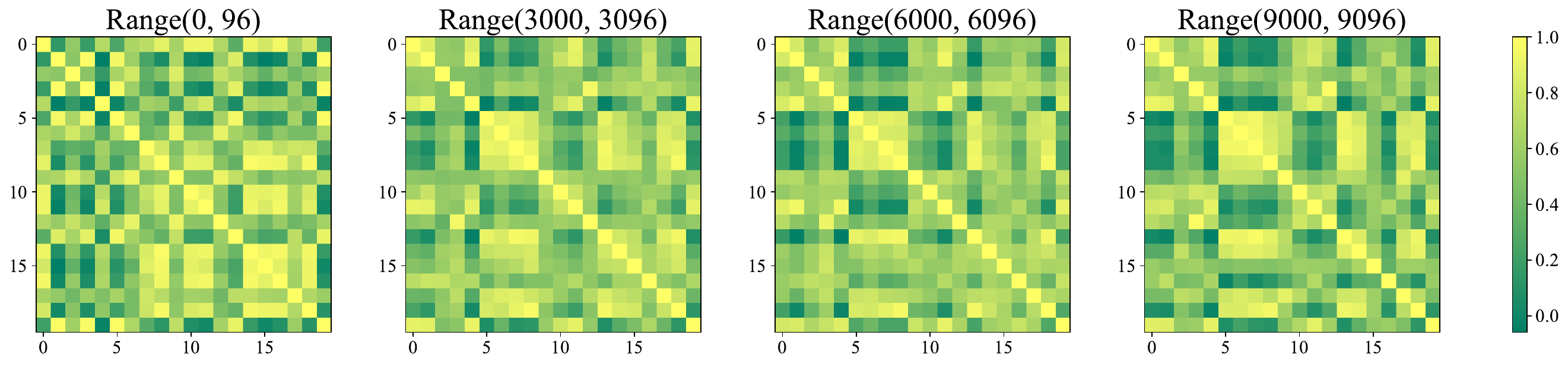}\\[-.2em]
        {\small (b) Correlation shifts in the Traffic dataset.}\\[.5em]
        \includegraphics[width=\linewidth]{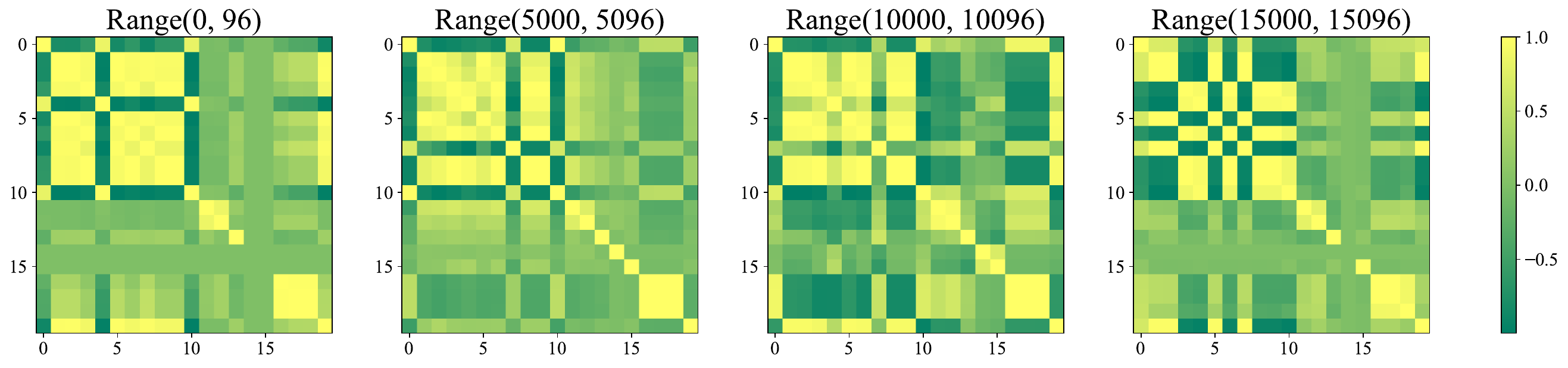}\\[-.2em]
        {\small (c) Correlation shifts in the Weather dataset.}\\[.5em]
        \includegraphics[width=\linewidth]{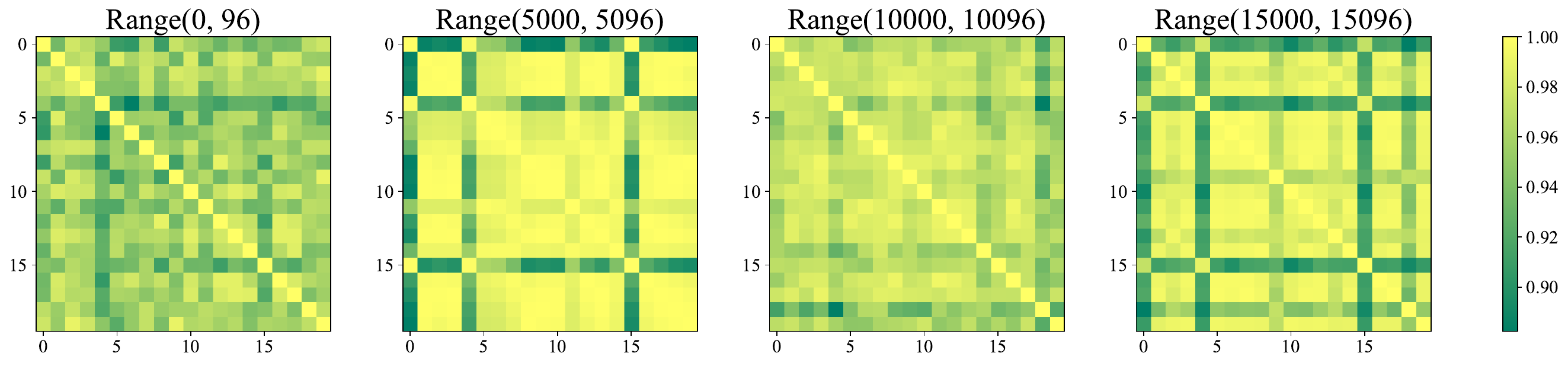}\\[-.2em]
        {\small (d) Correlation shifts in the Solar-Energy dataset.}\\[0em]
        \caption{Correlation matrix on the benchmark datasets across various ranges. The window length is set to 96. To aid visual understanding, only the initial 20 variates were utilized for each dataset. Range(\texttt{start}, \texttt{end}) indicates timestamps from \texttt{start} to \texttt{end}.}
        \label{fig:corr_matrix_varying_range}
        \vspace{-0.2cm}
    \end{figure}

    The correlation between variates can change as the sliding window moves through the data, resulting in unique correlations for each batch. The sliding window approach involves moving a fixed-size window over the time series data, capturing local temporal patterns and relationships among the variates within each window. Figure \ref{fig:corr_matrix_varying_range} illustrates correlation shifts among variates within diverse ranges of windows from each dataset. In all datasets, changes in the range of the target window lead to variations in the group of variates with strong correlations. These shifts occur due to the inherent property of time series data, where the distribution varies over time. The presence of correlation shifts in time series data necessitates the development of adaptive methods that can capture the dynamic \emph{batch-wise} relationships between variates.    

\newpage

\section{Proof of Theorem~\ref{theorem:error_k_DFH_approx}}
\label{appendix:proof_theorem_error_k_DFH_approx}
    \begin{proof}
        According to the Fourier transform, any time series $x(t)$ can be decomposed into a sum of sinusoidal signals:
        \begin{equation}
            x(t) = \sum_{i=1}^{n} A_i \sin(2 \pi f_i t + \phi_i)    
        \end{equation} 
        where $A_i$ is the amplitude, $f_i$ is the frequency, and $\phi_i$ is the phase of the $i$-th dominant sinusoidal component. Let $x_\text{recon}(t)$ be the reconstructed time series using only the $k$ dominant frequencies:
        \begin{equation}
            x_\text{recon}(t) = \sum_{i=1}^{k} A_i \sin(2 \pi f_i t + \phi_i)
        \end{equation}
        The error $\epsilon(t)$ can be defined as the difference between the original time series $x(t)$ and the reconstructed time series $x_\text{recon}(t)$:
        \begin{equation}
            \epsilon(t) = x(t) - x_\text{recon}(t) = \sum_{i=k+1}^{n} A_i \sin(2 \pi f_i t + \phi_i)
        \end{equation}
        The mean squared error (MSE) over the interval $[0, T]$ can be expressed as:
        \begin{equation}
            \begin{split}
                \text{MSE} = \frac{1}{T} \int_{0}^{T} \epsilon(t)^2 \, dt \\
                = \frac{1}{T} \int_{0}^{T} \left( \sum_{i=k+1}^{n} A_i \sin(2 \pi f_i t + \phi_i) \right)^2 \, dt
            \end{split}
        \end{equation}
        Using the orthogonality of sine functions, the cross terms in the integral will vanish, leaving us with:
        \begin{equation}
            \text{MSE} = \frac{1}{T} \sum_{i=k+1}^{n} A_i^2 \int_{0}^{T} \sin^2(2 \pi f_i t + \phi_i) \, dt
        \end{equation}
        Since $\sin^2(2 \pi f_i t + \phi_i)$ has an average value of $1/2$ over its period, the MSE simplifies to:
        \begin{equation}
            \text{MSE} = \frac{1}{T} \sum_{i=k+1}^{n} A_i^2 \cdot \frac{T}{2} = \frac{1}{2} \sum_{i=k+1}^{n} A_i^2
        \end{equation}
        
        Therefore, the error is given by the cumulative contribution of the squared amplitudes of the non-dominant frequencies. This ensures that as long as the $k$ dominant frequencies capture the majority of the signal's energy, the error will remain small.
    \end{proof}

\newpage
\section{Data Descriptions}
\label{sec:appendix:data_descriptions}
    Electricity contains the hourly electricity consumption of 321 customers from 2012 to 2014. Traffic consists of hourly road occupancy rates collected from 862 sensors located on San Francisco Bay Area freeways from 2015 to 2016. Weather consists of 21 meteorological indicators, such as air temperature, recorded every 10 minutes in 2020. Solar-Energy contains solar power production records from 2006, sampled every 10 minutes from 137 PV plants in Alabama. Following previous studies\cite{liu2023itransformer}, we partitioned the datasets into training, validation, and test sets in a 7:1:2 ratio chronologically. We also provide the detailed descriptions of the four benchmark datasets in Table~\ref{table:data_description}. 

    \begin{table}[h]
    
\centering
    \resizebox{\linewidth}{!}{
    \begin{tabular}{l|ccc}
        \toprule
        Dataset & \# Variates  &  Frequency & Size \\
        \midrule
        Electricity & 321 &  1 Hour & {\small(18317, 2633, 5261)}\\
        Trafic & 862&  1 Hour & {\small(12185, 1757, 3509})\\
        Weather & 21 &  10 min & {\small(36792, 5271, 10540})\\
        Solar-Energy & 137 &  10 min & {\small(36601, 5161, 10417})\\
        \bottomrule
    \end{tabular}
    }
    \caption{Details of datasets description. "Frequency" refers to the sampling interval of time points. The three numbers in "Size" represent the sizes of the train, validation, and test datasets, respectively.}
    
    \label{table:data_description}
\end{table}

\section{Implementation Details}
\label{sec:appendix:implementation_details}
    The overall experiment configuration of \algname{} follows the approach of popular previous works\,\cite{liu2023itransformer}. The batch size is set to 32 for all experiments. The settings for the two hyperparameters of \algname{} are presented in Table~\ref{table:hyperparameter_setting.tex}. The choice of these parameters depends on the requirements of the application and we selected them appropriately within the range of \(k \in \{3, 4\}\) and \(gs \in \{5, 10, 20\}\) without sophisticated tuning, leaving their detailed optimization for future work. For low-pass filtering, cut-off frequency $\varepsilon$ is set to 25, the half of the number of positive frequencies after Fourier transformation. Note that we did not perform any sophisticated tuning to determine these parameter values, and the choice of these parameters highly depends on the requirements of the application. The length of lookback window is set to 96 for all experiments. \algname{} is implemented using PyTorch 2.0.1 and tested on a single NVIDIA GeForce RTX 3090 24GB GPU.

    \begin{table}[h]
\small
    
\centering
    \begin{tabular}{l|ccc}
        \toprule

        Dataset      & $k$ & $gs$  & $\varepsilon$ \\
        \midrule
        Electricity  & 3 & 10 & 25\,(50\%) \\
        Traffic      & 4 & 10 & 25\,(50\%) \\
        Weather      & 4 &  5 & 25\,(50\%) \\ 
        Solar-Energy & 3 & 20 & 25\,(50\%) \\
        \bottomrule
    \end{tabular}

    \caption{Hyperparameter setting of \algname{}.}.

    \label{table:hyperparameter_setting.tex}
\end{table}

\newpage
\section{Further Experimental Results}
\label{appendix:further_experimental_results}
    Due to space limitations in the main text, we provide additional experimental results in this section. 
    
    \smallskip \noindent \textbf{Forecasting Performance with Standard Deviation} Table~\ref{table:standard_deviation_ours.tex} shows the average forecasting performance of \algname{}, along with the standard deviation across five independent runs. The low standard deviation values indicate that \algname{} consistently produces reliable results, demonstrating its robustness. 
    
    \def\arraystretch{1}
\begin{table}[h]
    
\setlength{\tabcolsep}{4pt}

\centering
\resizebox{\linewidth}{!}{
    \begin{tabular}{c|cc|cc}
    
        \toprule
        Dataset & \multicolumn{2}{c|}{Electricity} & \multicolumn{2}{c}{Traffic} \\[.4em]
        Metric & MSE & MAE & MSE & MAE \\
        \midrule
        96 & 0.153{$\pm$ 0.000} & 0.245{$\pm$ 0.000} & 0.396{$\pm$ 0.000} & 0.274{$\pm$ 0.000}\\
        192 & 0.167{$\pm$ 0.000} & 0.257{$\pm$ 0.001} & 0.417{$\pm$ 0.000} & 0.281{$\pm$ 0.000}\\
        336 & 0.183{$\pm$ 0.001} & 0.275{$\pm$ 0.001} & 0.435{$\pm$ 0.001} & 0.289{$\pm$ 0.000}\\
        720 & 0.220{$\pm$ 0.001} & 0.305{$\pm$ 0.001} & 0.472{$\pm$ 0.001} & 0.308{$\pm$ 0.000}\\
        \midrule
        Avg & 0.181{$\pm$ 0.001} & 0.271{$\pm$ 0.001} & 0.430{$\pm$ 0.001} & 0.288{$\pm$ 0.000}\\
        \bottomrule
        \addlinespace
        \toprule
        Dataset & \multicolumn{2}{c|}{Weather} & \multicolumn{2}{c}{Solar-Energy} \\[.4em]
        Metric & MSE & MAE & MSE & MAE \\
        \midrule
        96 & 0.178{$\pm$ 0.001} & 0.218{$\pm$ 0.001} & 0.202{$\pm$ 0.001} & 0.238{$\pm$ 0.001}\\
        192 & 0.227{$\pm$ 0.001} & 0.258{$\pm$ 0.000} & 0.237{$\pm$ 0.000} & 0.262{$\pm$ 0.000}\\
        336 & 0.281{$\pm$ 0.001} & 0.297{$\pm$ 0.001} & 0.254{$\pm$ 0.001} & 0.275{$\pm$ 0.001}\\
        720 & 0.357{$\pm$ 0.000} & 0.347{$\pm$ 0.000} & 0.252{$\pm$ 0.000} & 0.274{$\pm$ 0.000}\\
        \midrule
        Avg & 0.261{$\pm$ 0.001} & 0.280{$\pm$ 0.001} & 0.236{$\pm$ 0.001} & 0.262{$\pm$ 0.000}\\
        \bottomrule

    \end{tabular}
}

\caption{Multivariate time series forecasting results of \algname{} with averages and standard deviations. The experiment was conducted five times.}

\label{table:standard_deviation_ours.tex}
\end{table}

\section{Further Visualizations}
\label{appendix:further_analysis}
    
\begin{figure}[h]
        \centering
        \includegraphics[width=\linewidth]{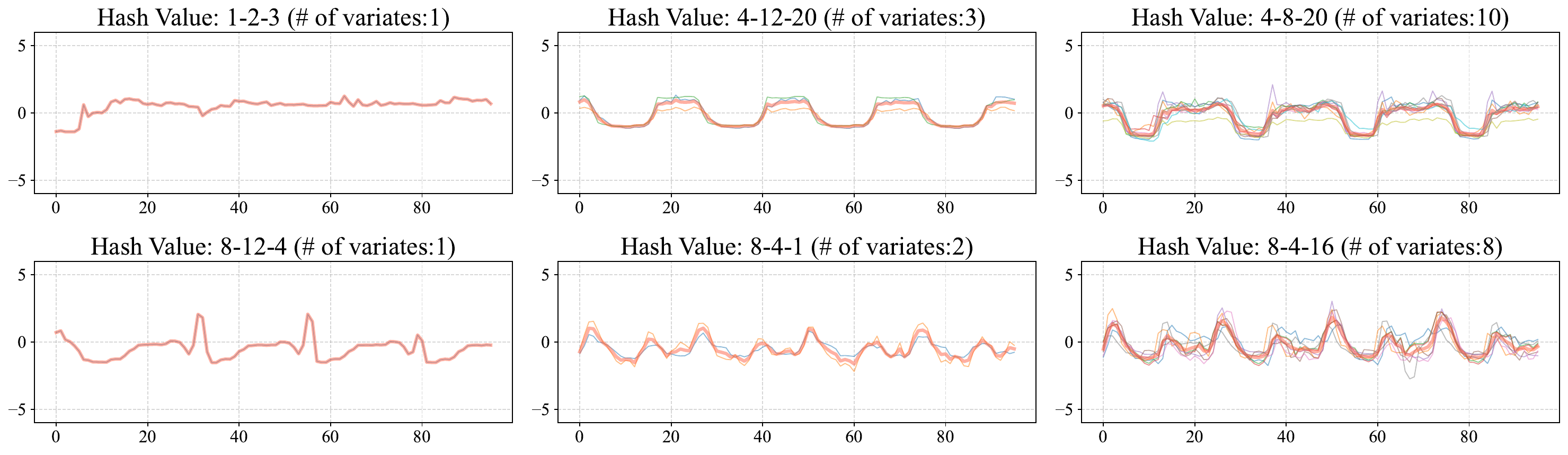}\\
        {\small (a) Electricity dataset\,($k$=3).}\\[.9em]
        \includegraphics[width=\linewidth]{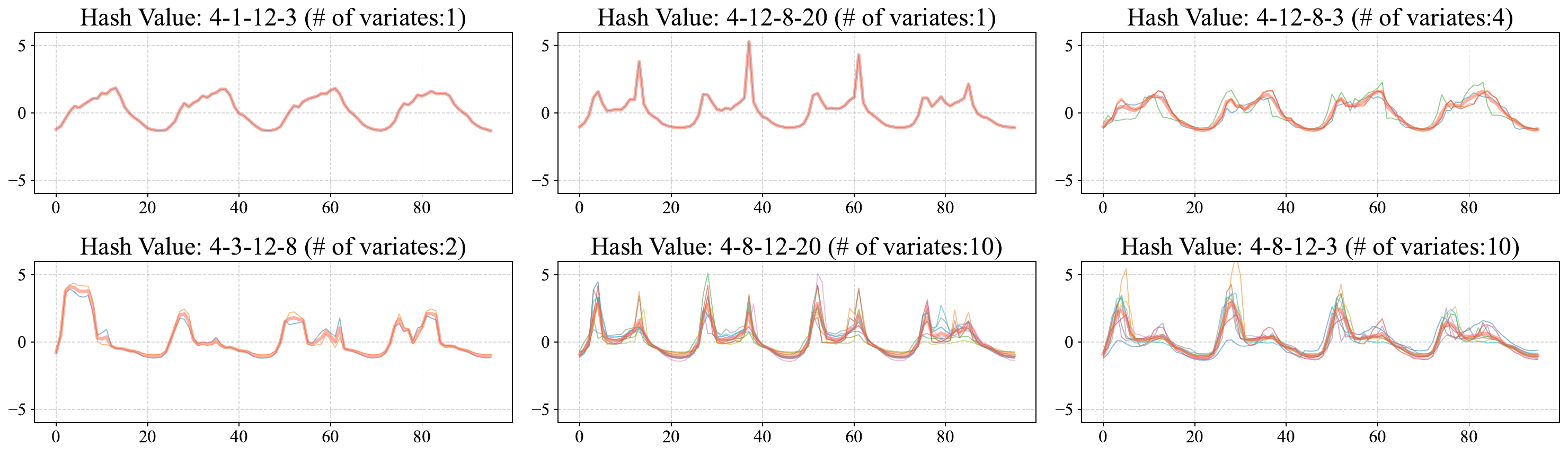}\\
        {\small (b) Traffic dataset\,($k$=4).}\\
        \caption{Additional visualizations of variates grouped by $k$-DFH. The red lines represent the mean values of the variates within each group, illustrating the distinctive periodic pattern of each group. The hash values represent the ordered set of $k$ dominant frequencies.}
        \label{fig:examples_hash_group}
    \end{figure}

    \textbf{Further Visualizations of k-DFH's Grouping Results.} We provide additional visualization of the groups generated by $k$-DFH in Figure~\ref{fig:examples_hash_group}. Each subfigure demonstrates how the variates are clustered based on their hash values, representing overall periodic behaviors. In this visualization, the order of dominant frequencies distinguishes different patterns. In Figure~\ref{fig:examples_hash_group}(b), two variates with hash values "4-8-12-20" and "4-12-8-20" display distinct periodic behaviors despite having the same set of dominant frequencies. In Figure~\ref{fig:examples_hash_group}(a), the first variate labeled "Hash Value: 1-2-3" displays a relatively flat trend, indicating minimal periodic behavior. This implies that the $k$-DFH method can effectively isolate variates that do not share common periodic characteristics with others.
    
\begin{figure}[ht]
        \centering
        \includegraphics[width=0.25\linewidth]{figures/correlations/corr_matrix_ECL.pdf}\hspace{.15cm}
        \includegraphics[width=0.25\linewidth]{figures/sparse_matrix/sparse_matrix_ECL_k3_gs10.pdf}\\
        {\small (a) Electricity dataset.}\\
        
        \includegraphics[width=0.25\linewidth]{figures/correlations/corr_matrix_Traffic.pdf}\hspace{.15cm}
        \includegraphics[width=0.25\linewidth]{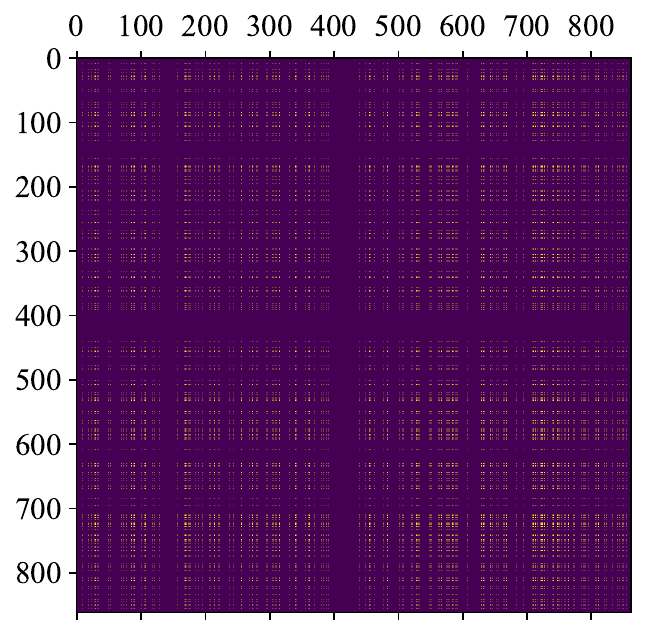}\\
        {\small (b) Traffic dataset.}\\
        
        \includegraphics[width=0.25\linewidth]{figures/correlations/corr_matrix_Weather.pdf}\hspace{.15cm}
        \includegraphics[width=0.25\linewidth]{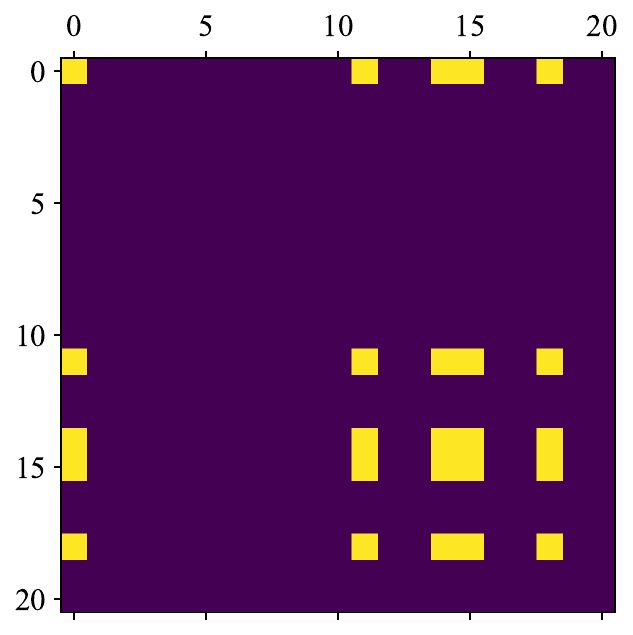}\\[-.2em]
        {\small (c) Weather dataset.}\\[-.5em]

        \caption{Comparison of correlation matrices and sparse matrices generated by \algname{}. Each subfigure includes the correlation matrix\,(left) and the corresponding 0-1 matrix from stratified sampling\,(right). Yellow indicates values close to 1, while darker colors represent values close to 0.}
        \label{fig:compare_corr_and_sparse_matrices}
    \end{figure}

    \textbf{Visualizations of Sparse Matrices.} Figure~\ref{fig:compare_corr_and_sparse_matrices} illustrates the correlation coefficients between variates in the last 96-length window of the training data and the corresponding sparse matrices generated by \algname{}. As clearly shown in the Electricity and Weather datasets, their sparse matrices reveal that most variates with low correlation to others are selected by \algname{}. This is because the variates exhibiting unique behaviors have dominant frequencies that differ from the others. The $k$-DFH algorithm identifies these as an independent group, allowing the selected variate tokens to effectively represent the most significant data distributions.

\end{document}